\documentclass[11pt]{article}

\usepackage[final]{acl}

\usepackage[T1]{fontenc}
\usepackage[utf8]{inputenc}
\usepackage{times}
\usepackage{latexsym}
\usepackage{microtype}
\usepackage{inconsolata}

\usepackage{amsmath}
\usepackage{amssymb}
\usepackage{amsfonts}
\usepackage{mathtools}

\usepackage{graphicx}
\usepackage{booktabs}
\usepackage{multirow}
\usepackage{subcaption}

\usepackage{xcolor}
\usepackage{url}
\usepackage{xurl}
\usepackage{hyperref}

\graphicspath{{figure/}}

\usepackage{adjustbox}
\usepackage{booktabs}
\usepackage{multirow}

\title{Do LLMs Trust the Accuser or the Accusation? Measuring Belief Shifts in Werewolf}

\makeatletter
\def\@fnsymbol#1{%
  \ifcase#1
  \or \textdagger
  \or \textasteriskcentered
  \or \textdaggerdbl
  \or \S
  \or \P
  \or \ensuremath{\parallel}
  \or \textasteriskcentered\textasteriskcentered
  \or \textdagger\textdagger
  \or \textdaggerdbl\textdaggerdbl
  \else
    \@ctrerr
  \fi
}
\makeatother

\author{
  \textbf{Yu-Yu Yang\textsuperscript{1, 2}},
  \textbf{Ti-Rong Wu\textsuperscript{2}\thanks{Corresponding author.}},
  \textbf{Hung Guei\textsuperscript{2, 3}},
  \textbf{Hsing-Yu Chen\textsuperscript{1, 2}},
  \textbf{I-Chen Wu\textsuperscript{1, 4}}\\
  \vspace{1pt}\\
  \textsuperscript{1}Department of Computer Science, National Yang Ming Chiao Tung University, Taiwan \\
  \textsuperscript{2}Institute of Information Science, Academia Sinica, Taiwan\\
  \textsuperscript{3}College of Artificial Intelligence, National Yang Ming Chiao Tung University, Taiwan\\
  \textsuperscript{4}Research Center for Information Technology Innovation, Academia Sinica, Taiwan\\
  \texttt{yyyang.cs13@nycu.edu.tw, tirongwu@iis.sinica.edu.tw}\\
}

\begin{document}
\maketitle

\begin{abstract}
Social-deduction games such as Werewolf are increasingly used to evaluate LLM agents, but existing evaluations often rely on final game outcomes.
We propose a belief-shift evaluation benchmark in Werewolf for analyzing communication skills through belief updating.
Using LLM-played games, we annotate suspicion and accusation messages and measure how an observing village-side model's beliefs change after each message.
We evaluate 40 open-weight LLM configurations on 1,224 annotated messages.
Our results show that larger models better distinguish true wolves from villagers based on game history, but accusations still strongly influence their beliefs.
Models become more suspicious of the accused target and less suspicious of the accuser, especially when the accuser is trusted, even if the accuser is wolf-aligned.
Larger models better resist accusations from accusers they already distrust.
Overall, our findings suggest that current open-weight LLMs up to 120B parameters still struggle to integrate accusation content with source trust in strategic communication.
Our benchmark and code are available at \url{https://rlg.iis.sinica.edu.tw/papers/werewolf-accusation-benchmark}.
\end{abstract}

\section{Introduction}
\label{sec:intro}

Large language models (LLMs) are increasingly used as agents~\cite{park_generative_2023,wu_autogen_2024,li_camel_2023} that communicate with humans and other agents.
In such settings, success depends not only on solving a task, but also on an agent's communication skills, including the ability to express itself and understand others.
This makes communication a key ability for agentic AI.
However, most static benchmarks~\cite{phan_benchmark_2026,wang_mmlupro_2024} are not well-suited to evaluate this ability, since they usually present fixed inputs and require fixed answers without interaction.

Recent studies have therefore moved toward interactive evaluations, including multi-agent strategic communication games such as Werewolf~\cite{xu_exploring_2024,xu_language_2024}, where agents must reason about hidden information, persuade others, and deceive opponents during interaction.
Although these games provide suitable interactive benchmarks, current evaluations often rely on final game outcomes, such as win/loss or win rate.
Such metrics are straightforward, but they provide little insight into the communication behaviors that lead to success or failure.
This raises a central challenge: \textit{how to measure the communication behaviors of LLM agents beyond final game outcomes?}

To address this challenge, this paper focuses on \textit{belief change} as an intermediate measure of communication behavior.
Here, \textit{belief} refers to an agent's subjective judgment about hidden information, such as which players are likely to be allies, opponents, or deceivers.
In Werewolf, messages often aim to influence such judgments, for example by making other agents accept a particular viewpoint.
Belief change therefore provides a direct way to analyze whether an agent accepts, rejects, or discounts information from others.
This perspective is closely related to theory of mind~\cite{premack_does_1978,baron-cohen_does_1985}, since appropriate belief updating requires reasoning about the knowledge, intention, and reliability behind a message.

Based on this perspective, we construct a belief-shift dataset from LLM-played Werewolf games by annotating specific messages in which one agent accuses another player.
For each message, we measure how an observing agent's belief about both the accuser and the accused target changes from before to after the message.
This allows us to analyze how the message changes agents' beliefs during interaction, rather than using the final game outcome.

Using this dataset, we evaluate 40 open-weight LLM configurations from major model families.
Our results reveal three findings.
First, models show some ability to distinguish true wolves from true villagers, and this ability improves with scale.
Second, when models already trust the accuser, they tend to accept the accusation without considering that the accuser may be mistaken or deceptive.
Third, even when models do not trust the accuser, they still show a tendency to shift their beliefs toward the accusation.
Overall, these results suggest that current open-weight LLMs up to 120B parameters still lack a reliable ability to integrate message content with source trust when updating beliefs in strategic communication.
We also release the annotated game logs and the evaluation code, so that new models can be evaluated on the same messages under the same protocol.

\section{Related Work}
\label{sec:related}

\subsection{Belief Reasoning and Persuasion in LLMs}
\label{sec:related-belief-updating}

Most existing work on theory of mind (ToM) and belief reasoning evaluates LLMs through fixed narratives or QA benchmarks, such as ToMi \cite{le_revisiting_2019}, BigToM \cite{gandhi_understanding_2023}, FANToM \cite{kim_fantom_2023}, and ToMBench \cite{chen_tombench_2024}.
These benchmarks test whether a model can infer a character's belief from a fixed context, whereas we measure how the model's own beliefs change during an interactive multi-agent game.

Relatedly, several studies~\cite{durmus2024persuasion,xu_earth_2024,tan_persuasion_2025,bajaj_who_2026} investigate LLM persuasion and susceptibility to persuasion: \citet{durmus2024persuasion} show that LLMs can produce persuasive arguments, while \citet{xu_earth_2024} show that LLMs' beliefs can be shifted through multi-turn persuasion.
\citet{bajaj_who_2026} further show that LLMs are more likely to accept claims from sources that claim to be experts, even when those claims are incorrect.
Our work instead studies observer-side belief updating in a multi-agent strategic game, where an evaluated model observes one player accusing another and reports belief shifts about both players.

\subsection{LLMs in Social-Deduction Games}
\label{sec:related-social-deduction}

Social-deduction games such as Werewolf~\cite{xu_learning_2025,xu_language_2024} and Avalon~\cite{light_avalonbench_2023} have been widely used to evaluate LLM agents because they require hidden-role reasoning, communication, persuasion, and deception.
Prior work~\cite{xu_learning_2025,xu_language_2024} mainly focuses on training stronger game-playing agents and evaluates agents through game-level outcomes, such as win/loss.
Closest to our setting, WOLF~\cite{agarwal_wolf_2025} focuses on deception generation and detection at the utterance level.
For each statement, listeners predict a deception category, and its main evaluation treats each utterance as a classification problem.
Our work instead takes a theory-of-mind perspective and examines how an accusation changes the observer's belief about both the accuser and the accused target.
Our benchmark thus extends WOLF from detection accuracy to belief change.

\section{Belief-Shift Measurement in Werewolf}
\label{sec:framework}

\begin{figure*}[t]
  \centering
  \includegraphics[width=\linewidth]{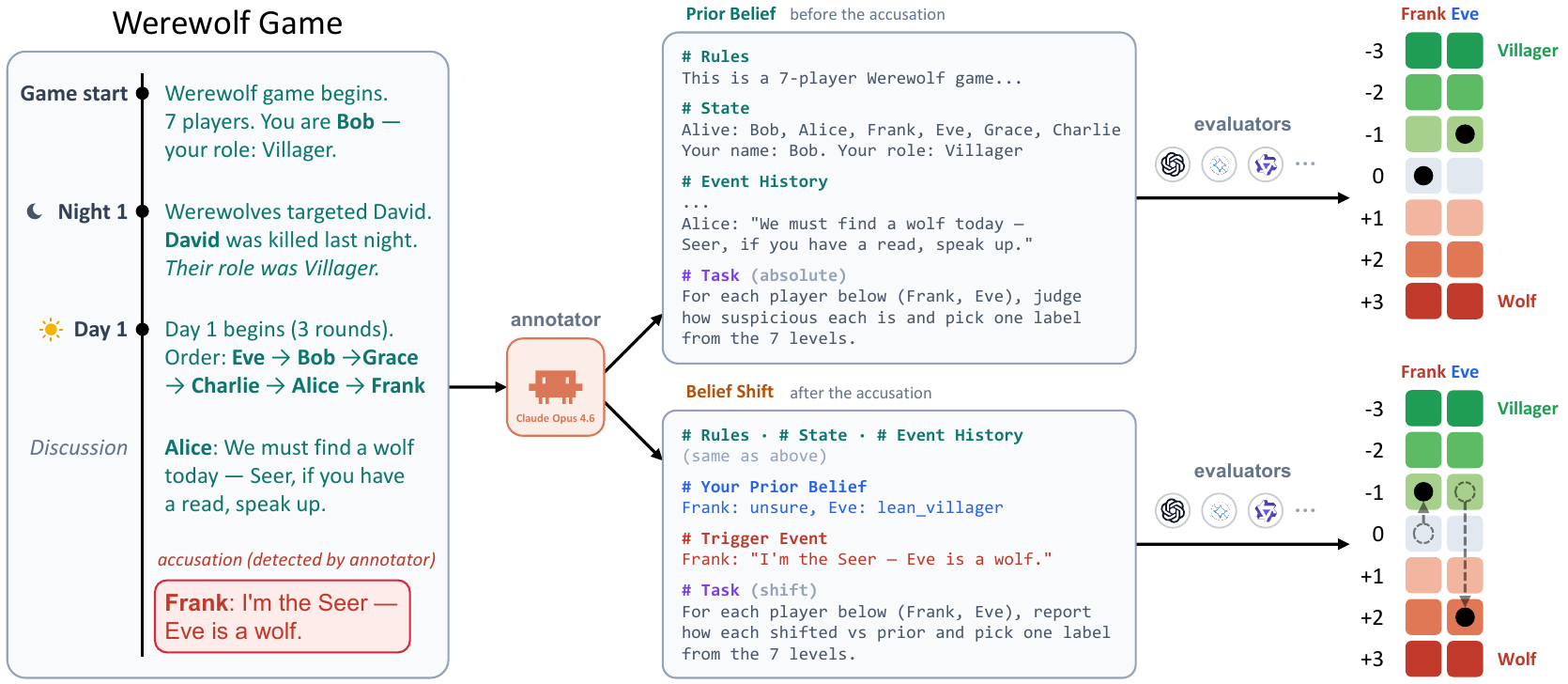}
  \caption{%
    Overview of belief-shift measurement pipeline.
    For each game, we annotate accusation messages and reconstruct an observing agent's view before and after each message.
    The model first reports its prior belief about the accuser and the accused target, and then reports the belief shift caused by the message.
  }
  \label{fig:pipeline}
\end{figure*}

Figure~\ref{fig:pipeline} introduces the procedure for measuring belief shifts in Werewolf, including two steps: generating a dataset from LLM-played Werewolf games and annotating accusation messages, followed by measuring how observing agents' beliefs change before and after each message.

\subsection{Dataset Generation}
\label{sec:framework-data}

We first generate 200 seven-player Werewolf games played by LLM agents.
For each game, each agent is randomly assigned a role.
During gameplay, each agent receives the game rules and the event history observable to its role, and is prompted to produce the next legal action, without any strategic guidance~\cite{xu_language_2024,xu_learning_2025}.
After each game is completed, we annotate the daytime discussion to identify messages in which a player either \textit{suspects} another player of being on the wolf team, i.e., expresses uncertainty or raises the possibility that the player may be wolf-aligned, or \textit{accuses} another player, i.e., directly states that the player is on the wolf team\footnote{We use Claude Opus 4.6 as the annotator, and human annotators verify a subset of the annotations.}.
For each marked message, we record the \textit{accuser}, the \textit{accused target}, and their roles and teams.
Both the accuser and the accused target may belong to either the village team or the wolf team.
Overall, we obtain 1,224 annotated messages for belief-shift evaluation.

\subsection{Measuring Belief Shifts}
\label{sec:framework-measuring}

For each annotated message, we measure belief shifts from the perspective of one or two randomly selected surviving village-team players.
Before the message, the evaluated model receives the game history, including its role and all events observable to that role.
We then ask the model to report its belief about both the accuser and the accused target on a seven-level scale, ranging from $-3$ to $+3$.
Here, $-3$ indicates that the player is believed to be strongly village-side, while $+3$ indicates that the player is believed to be strongly wolf-side.

After the message, the evaluated model receives the same game history, together with two additional inputs: the accusation message and its prior belief from the before-message stage.
We then ask the model to report the belief shift for both the accuser and the accused target.
The belief shift is also represented on a seven-level scale from $-3$ to $+3$, where positive values indicate a shift toward the wolf side and negative values indicate a shift toward the village side.

We evaluate 40 open-weight LLM configurations from major model families, ranging from 1B to 120B parameters.
The full model list is provided in Appendix~\ref{app:models-list}.

\section{Results}
\label{sec:results}

This section presents the results.
We test each scale-dependent trend in each experiment with a Spearman rank correlation between model size and the reported quantity.
95\% confidence intervals for all reported means are provided in Appendix~\ref{app:result-reporting}.

\subsection{Prior Belief Discrimination}
\label{sec:results-prior}

\begin{figure}[t]
  \centering
  \includegraphics[width=\linewidth]{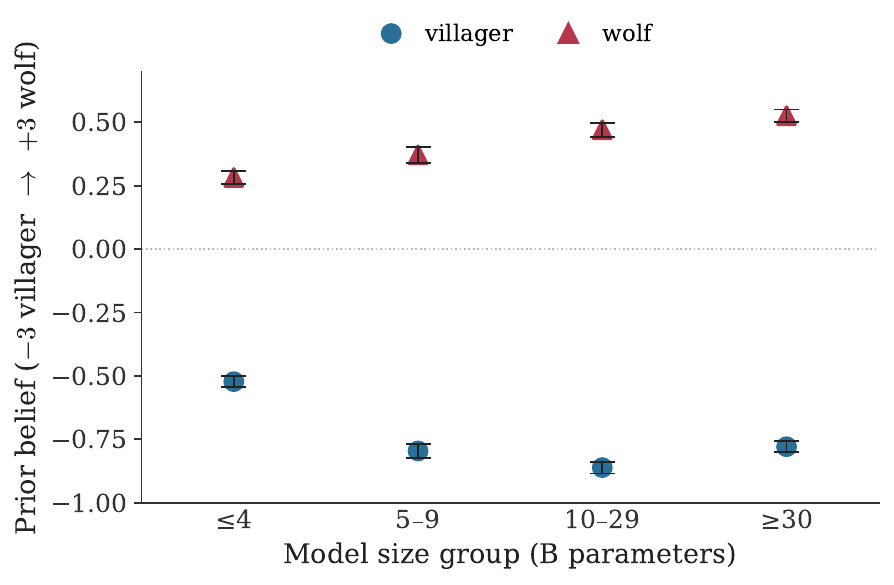}
  \caption{Prior beliefs by model size group, on the $-3$ to $+3$ wolf-positive scale ($-3$ = villager, $+3$ = wolf). Error bars show 95\% confidence intervals.}
  \label{fig:prior_belief}
\end{figure}

We first examine whether models can produce accurate prior beliefs by identifying wolves and villagers from the game context.
Specifically, before observing the annotated message, each model is asked to estimate how likely another player is to be on the wolf team.
Figure~\ref{fig:prior_belief} shows the average prior belief assigned to true villagers and true wolves across different model sizes.

As shown in the figure, models already show some ability to discriminate hidden roles from the game context.
Across all size groups, true wolves receive higher wolf-leaning prior beliefs than true villagers.
This separation generally becomes larger with model size, increasing from 0.80 in the $\le$4B group to around 1.3 for models above 10B (Spearman $\rho=0.54$, $p=4\times10^{-4}$), suggesting that larger models are better at identifying hidden roles from the observed game history.

\subsection{Belief Shifts after Accusations}
\label{sec:results-shift}

\begin{table}[t]
  \centering
  \footnotesize
  \setlength{\tabcolsep}{4.5pt}
  \renewcommand{\arraystretch}{0.95}
  \begin{tabular}{l r rr rr}
    \toprule
    & & \multicolumn{2}{c}{Target} & \multicolumn{2}{c}{Accuser} \\
    \cmidrule(lr){3-4} \cmidrule(lr){5-6}
    Size group & $n$ & Sus. & Acc. & Sus. & Acc. \\
    \midrule
    \multicolumn{6}{l}{\textbf{Villager as accuser}} \\
    $\le$4B & 12 & 0.24 & 0.94 & -0.63 & -0.54 \\
    5--9B & 8 & 0.52 & 1.03 & -0.87 & -0.75 \\
    10--29B & 9 & 0.64 & 1.04 & -0.37 & -0.36 \\
    $\ge$30B & 11 & 0.62 & 0.96 & -0.47 & -0.44 \\
    \cmidrule(lr){1-6}
    Avg. & 40 & 0.49 & 0.98 & -0.57 & -0.51 \\
    \midrule
    \multicolumn{6}{l}{\textbf{Wolf as accuser}} \\
    $\le$4B & 12 & 0.21 & 0.70 & -0.64 & -0.47 \\
    5--9B & 8 & 0.33 & 0.88 & -0.77 & -0.73 \\
    10--29B & 9 & 0.35 & 0.78 & -0.09 & -0.14 \\
    $\ge$30B & 11 & 0.34 & 0.66 & -0.13 & -0.08 \\
    \cmidrule(lr){1-6}
    Avg. & 40 & 0.30 & 0.74 & -0.40 & -0.34 \\
    \bottomrule
  \end{tabular}
  \caption{%
    Average belief shifts after suspicion or accusation messages, grouped by the accuser's true team and message strength.
    \textbf{Sus.} denotes a soft suspicion, and \textbf{Acc.} denotes a direct accusation.
    $n$ is the number of models in each size group.
    95\% confidence intervals for these means are provided in Table~\ref{tab:exp2_belief_shift_stacked_rowlevel_with_ci}.
  }
  \label{tab:exp2_belief_shift_stacked}
\end{table}

Next, we examine how accusation messages change models' beliefs about both the accuser and the accused target.
Table~\ref{tab:exp2_belief_shift_stacked} reports the average belief shifts grouped by the accuser's true team and the message strength.
Positive values indicate a shift toward the wolf side, while negative values indicate a shift toward the village side.

Interestingly, after observing a suspicion or accusation message, models generally become more suspicious of the accused target and less suspicious of the accuser.
Direct accusations (Acc.) also lead to stronger belief shifts than soft suspicions (Sus.).
For example, the target shift increases from 0.49 under soft suspicion to 0.98 under direct accusation when the accuser is on the village team, and from 0.30 to 0.74 when the accuser is on the wolf team.

In addition, models show only limited sensitivity to whether the accuser is in fact on the wolf team.
Although such accusations are often strategically misleading in Werewolf, the accused target still receives a positive shift across all model sizes.
This suggests that models remain influenced by the accusation itself even when the source may be unreliable.
However, larger models are less likely to treat the accuser as reliable: under direct accusations, the accuser shift changes from $-0.47$ in the $\leq$4B group to $-0.08$ in the $\geq$30B group (Spearman $\rho=0.44$, $p=0.004$), although the accused target still shifts toward the wolf side.

\subsection{Belief Shifts by Prior Trust}
\label{sec:results-prior-trust}

\begin{table}[t]
  \centering
  \scriptsize
  \setlength{\tabcolsep}{3pt}
  \renewcommand{\arraystretch}{0.95}
  \begin{tabular}{l r ccc ccc}
    \toprule
    & & \multicolumn{3}{c}{Target} & \multicolumn{3}{c}{Accuser} \\
    \cmidrule(lr){3-5} \cmidrule(lr){6-8}
    Size & $n$ & Trust & Neutral & Distrust & Trust & Neutral & Distrust \\
    \midrule
    \multicolumn{8}{l}{\textbf{Villager as accuser}} \\
    $\le$4B & 12 & 0.78 & 0.72 & 0.37 & -0.55 & -0.46 & -0.97 \\
    5--9B & 8 & 1.04 & 0.47 & 0.49 & -0.75 & -0.84 & -0.99 \\
    10--29B & 9 & 0.99 & 0.82 & 0.20 & -0.35 & -0.48 & 0.11 \\
    $\ge$30B & 11 & 0.96 & 0.72 & 0.10 & -0.43 & -0.63 & 0.39 \\
    \cmidrule(lr){1-8}
    Avg. & 40 & 0.94 & 0.71 & 0.33 & -0.51 & -0.57 & -0.56 \\
    \midrule
    \multicolumn{8}{l}{\textbf{Wolf as accuser}} \\
    $\le$4B & 12 & 0.72 & 0.50 & 0.30 & -0.58 & -0.48 & -0.56 \\
    5--9B & 8 & 1.18 & 0.37 & 0.26 & -0.73 & -0.80 & -0.71 \\
    10--29B & 9 & 0.97 & 0.62 & 0.15 & -0.32 & -0.17 & 0.18 \\
    $\ge$30B & 11 & 1.00 & 0.60 & -0.12 & -0.43 & -0.25 & 0.47 \\
    \cmidrule(lr){1-8}
    Avg. & 40 & 0.96 & 0.53 & 0.14 & -0.52 & -0.40 & -0.14 \\
    \bottomrule
  \end{tabular}
  \caption{%
    Average belief shifts under different levels of prior trust in the accuser.
    Rows are grouped by the accuser's true team and model-size group.
    \textbf{Trust} ($\{-3,-2\}$), \textbf{Neutral} ($\{-1,0,+1\}$), and \textbf{Distrust} ($\{+2,+3\}$) are defined by the observer's prior belief about the accuser.
    $n$ is the number of models in each size group.
    95\% confidence intervals for these means are provided in Table~\ref{tab:exp3_credulity_by_prior_by_size_stacked_rowlevel_with_ci}.
  }
  \label{tab:exp3_credulity_by_prior_by_size_stacked_rowlevel}
\end{table}

Finally, we examine whether belief shifts depend on the model's prior trust in the accuser.
Since the observer is always on the village team, a village-leaning prior belief about the accuser indicates higher trust, while a wolf-leaning prior belief indicates lower trust.
We divide examples into three bins according to the prior belief about the accuser: \textit{Trust}, \textit{Neutral}, and \textit{Distrust}.
Table~\ref{tab:exp3_credulity_by_prior_by_size_stacked_rowlevel} reports the belief shifts under these trust levels.

When the accuser is trusted, models show the strongest shift toward the accusation.
The accused target receives a large positive shift regardless of whether the accuser is actually on the village team or the wolf team (0.94 vs. 0.96).
This suggests that once models already trust the accuser, they tend to accept the accusation even when the accuser is in fact wolf-aligned.

When the accuser is distrusted, the shifts become smaller but do not disappear.
On average, the accused target still shifts toward the wolf side, while the accuser still shifts toward the village side.
This suggests that even when models distrust the accuser, their beliefs still move slightly in the direction suggested by the accusation.
However, larger models show a stronger ability to resist such accusations: for wolf-side accusers in the Distrust bin, the target shift decreases from 0.30 in the $\leq$4B group to $-0.12$ in the $\geq$30B group (Spearman $\rho=-0.37$, $p=0.018$), while the accuser shift changes from $-0.56$ to 0.47 (Spearman $\rho=0.63$, $p<10^{-4}$).
These results suggest that models are most susceptible to trusted accusers, while larger models make better use of prior distrust when the accuser is already viewed as unreliable.

\section{Discussion}
\label{sec:discussion}

This paper presents a belief-shift evaluation framework for analyzing communication behavior in LLM agents beyond final game outcomes.
Using LLM-played Werewolf games, we measure how an observing model updates its beliefs about both the accuser and the accused target before and after accusation messages.
Our results show that larger models are better at discriminating hidden roles from the game history, but they still tend to accept accusations.
After accusations, models generally shift suspicion toward the accused target and away from the accuser; this tendency is stronger when the accuser is trusted, even when the accuser is wolf-aligned.

Future directions include exploring broader communication skills beyond accusations, such as defense, questioning, and role claims, and extending the method to other social-deduction games.
Another direction is to extend the framework to speaker-side analysis, examining whether stronger models generate more strategic accusations and how effectively they influence different receivers. 
It would also be interesting to study how explicit strategic instructions, such as prompting agents to be more skeptical, affect the observed belief-shift patterns.
Furthermore, our current design uses pre-collected game logs to provide a unified evaluation setting in which all models receive the same contexts.
This design could also be extended to broaden the dataset, for example by incorporating mixed human--LLM games to examine whether similar patterns arise with human participants.

\section*{Limitations}

First, our evaluation covers only open-weight models (1B--120B), and does not include frontier closed-source systems such as GPT, Claude, and Gemini. The scale-dependent patterns we report, especially the regime change around 10B, should therefore be re-examined on those systems before being taken as universal.

Second, all observations come from a single environment, Werewolf. Its turn-based structure and its single-objective role of identifying wolves make accusations dense and salient, which is what lets us measure cleanly at the level of individual utterances. The same property may also amplify the credulity pattern relative to other interactive settings. Settings not covered here include negotiation, where accusations are rarer but stakes are higher; debate, where the audience is a third party rather than a peer; and cooperative tasks, where trust is positive-sum rather than zero-sum.

Third, we measure only the receiver side of belief updating. We do not characterise how the same model behaves as a speaker producing accusations. A complete picture of communication skill will require pairing this work with a speaker-side counterpart.

Fourth, as our goal is to measure whether an accusation changes the observer's belief relative to its belief immediately before the message, we provide the model with its own prior belief.
Since most of the context remains unchanged, this encourages the model to focus on the effect of the accusation itself.
However, providing the prior belief may introduce an anchoring effect.
An alternative is to elicit the prior and posterior beliefs independently.
Although the resulting prompts would differ only by the final accusation, the model may reassess the entire context rather than isolate the effect of the accusation.
We therefore leave a direct comparison of these two settings to future work.

Overall, our framework measures one specific receiver-side skill (belief updating around accusations) within one controlled environment. It is not intended as a comprehensive evaluation of LLM communication skill. It is a diagnostic instrument that isolates the source-versus-content trade-off in listener belief updating, and the same paired pre/post methodology can be extended to other receiver-side sub-skills.

\section*{Ethical Considerations}

\paragraph{Potential risks.}
Our findings characterise a failure mode, uncritical credulity toward trusted speakers, that could in principle be exploited adversarially. One concrete example is to first establish trust with an LLM agent and then inject misleading content into its pipeline. We surface this pattern in the hope that diagnostic visibility makes it easier to defend against rather than easier to abuse.

\paragraph{Annotation quality and annotator bias.}
Our annotation step uses a single LLM (Claude Opus 4.6) to mark suspect/accuse utterances. To verify annotation reliability, we drew a stratified subset of 150 utterances; two of the authors independently re-labelled this subset, and Claude Opus 4.6's labels agreed with the human consensus on 144 of 150 items (96\%). Residual annotator bias may still propagate into our category statistics, but the verified agreement bounds its expected magnitude.

\paragraph{Computational cost.}
All experiments are inference-only; we do not train or fine-tune any model. We evaluate 40 open-weight models ranging from 1B to 120B parameters (full list in Appendix~\ref{app:models}) on a workstation with four NVIDIA RTX Pro 6000 Blackwell GPUs. The complete evaluation, including self-play game collection, belief elicitation, and ablations, took approximately 72 wall-clock hours, corresponding to about 288 GPU-hours in total.

\paragraph{Dataset content and human subjects.}
This study involves no human participants. All game dialogue is generated by LLM self-play, and all belief responses are produced by the evaluated models themselves. The only annotator in the pipeline is Claude Opus 4.6, used to mark suspect/accuse utterances as described above. Because the corpus is entirely LLM-generated, it contains no human personally-identifying information and requires no anonymisation step. The corpus does contain hostile and deceptive utterances such as accusations, threats, and strategic lies. These are inherent to the social-deduction setting and are the object of study. No real-world targets are named.

\paragraph{Use of AI assistants.}
Beyond the annotation role disclosed above, AI assistants (large language models such as Claude and ChatGPT) were used during manuscript preparation and implementation: for English language polishing of author-drafted text; for generating LaTeX, table, and figure code; and for drafting parts of the experimental code. All AI-generated code was reviewed by the authors before being used in the reported experiments, and all resulting experimental outputs were manually inspected for correctness before being included in the paper. All research design, experimental decisions, analyses, claims, and interpretations are the authors' own.

\section*{Acknowledgements}
This research is partially supported by the National Science and Technology Council (NSTC) of the Republic of China (Taiwan) under Grant Number NSTC 114-2634-F-A49-004, NSTC 113-2221-E-001-009-MY3, NSTC 115-2221-E-A49-001, and NSTC 115-2221-E-A49-002. The authors would also like to thank the anonymous reviewers for their valuable comments.

\bibliography{reference_camera_ready}

\appendix
\section{Evaluated Models}
\label{app:models}

\subsection{Model List}
\label{app:models-list}

\begin{table*}[t]
  \centering
  \footnotesize
  \setlength{\tabcolsep}{4pt}
  \renewcommand{\arraystretch}{0.95}
  \begin{tabular}{rllrl}
    \toprule
    \# & Family & Model ID & Size (B) & Reasoning \\
    \midrule
    1  & GLM      & \texttt{zai-org/GLM-4.7-Flash}                   & 30   & on   \\
    2  & GPT-OSS  & \texttt{openai/gpt-oss-20b}                      & 20   & both \\
    3  & GPT-OSS  & \texttt{openai/gpt-oss-120b}                     & 120  & both \\
    4  & Gemma    & \texttt{google/gemma-4-E2B-it}                   & 2    & both \\
    5  & Gemma    & \texttt{google/gemma-4-E4B-it}                   & 4    & both \\
    6  & Gemma    & \texttt{google/gemma-4-26B-A4B-it}               & 26   & both \\
    7  & Gemma    & \texttt{google/gemma-4-31B-it}                   & 31   & both \\
    8  & Llama    & \texttt{meta-llama/Llama-3.2-1B-Instruct}        & 1    & off  \\
    9  & Llama    & \texttt{meta-llama/Llama-3.2-3B-Instruct}        & 3    & off  \\
    10 & Llama    & \texttt{meta-llama/Llama-3.1-8B-Instruct}        & 8    & off  \\
    11 & Llama    & \texttt{meta-llama/Llama-3.3-70B-Instruct}       & 70   & off  \\
    12 & Mistral  & \texttt{mistralai/Ministral-3-3B-Instruct-2512}  & 3    & off  \\
    13 & Mistral  & \texttt{mistralai/Ministral-3-8B-Instruct-2512}  & 8    & off  \\
    14 & Mistral  & \texttt{mistralai/Ministral-3-14B-Instruct-2512} & 14   & off  \\
    15 & Nemotron & \texttt{nvidia/NVIDIA-Nemotron-Nano-9B-v2}       & 9    & on   \\
    16 & Olmo     & \texttt{allenai/Olmo-3-7B-Instruct}              & 7    & off  \\
    17 & Olmo     & \texttt{allenai/Olmo-3-7B-Think}                 & 7    & on   \\
    18 & Olmo     & \texttt{allenai/Olmo-3.1-32B-Instruct}           & 32   & off  \\
    19 & Olmo     & \texttt{allenai/Olmo-3.1-32B-Think}              & 32   & on   \\
    20 & Phi      & \texttt{microsoft/Phi-4-mini-instruct}           & 3.8  & off  \\
    21 & Phi      & \texttt{microsoft/phi-4}                         & 14   & off  \\
    22 & Qwen     & \texttt{Qwen/Qwen3.5-2B}                         & 2    & both \\
    23 & Qwen     & \texttt{Qwen/Qwen3.5-4B}                         & 4    & both \\
    24 & Qwen     & \texttt{Qwen/Qwen3-8B}                           & 8    & on   \\
    25 & Qwen     & \texttt{Qwen/Qwen3.5-9B}                         & 9    & both \\
    26 & Qwen     & \texttt{Qwen/Qwen3-14B}                          & 14   & on   \\
    27 & Qwen     & \texttt{Qwen/Qwen3.6-27B}                        & 27   & both \\
    28 & Qwen     & \texttt{Qwen/Qwen3-30B-A3B}                      & 30   & on   \\
    29 & Qwen     & \texttt{Qwen/Qwen3.6-35B-A3B}                    & 35   & both \\
    \bottomrule
  \end{tabular}
  \caption{The 29 open-weight checkpoints evaluated in this paper, totalling 40 (checkpoint, reasoning-mode) configurations. The \emph{Reasoning} column shows whether reasoning is enabled: \emph{on} = reasoning-only, \emph{off} = no reasoning, \emph{both} = both settings evaluated as separate configurations. For Qwen, GPT-OSS, and Gemma~4, the reasoning toggle is controlled at inference time via a flag (e.g.,~\texttt{enable\_thinking}) rather than by separate checkpoints; for Olmo, \texttt{Instruct} and \texttt{Think} are released as distinct checkpoints.}
  \label{tab:models}
\end{table*}

Table~\ref{tab:models} lists the 29 open-weight checkpoints used in our evaluation, totalling 40 (checkpoint, reasoning-mode) configurations.

\subsection{Licenses and Intended Use}
\label{app:models-licenses}

All evaluated models are released under permissive licenses that allow academic research.
Qwen~3 \cite{yang_qwen3_2025}, GPT-OSS \cite{openai_gptoss120b_2025}, Gemma~4 \cite{_gemma_}, Mistral/Ministral~3 \cite{liu2026ministral3}, and Olmo~3 \cite{olmo_olmo_2025} are released under the Apache~2.0 license.
GLM-4.7-Flash \cite{team_glm45_2025} is released under the MIT license on Hugging Face. We follow this since our use depends on the HF weights; the GitHub codebase is Apache~2.0, which does not affect weight redistribution.
The Phi-4 family \cite{abdin2024phi4technicalreport,microsoft_phi4mini_2025} is also released under the MIT license.
Llama~3 models \cite{grattafiori_llama_2024} are released under the Llama~3 Community License, which permits academic use; the 700M monthly-active-user threshold and the ``Built with Llama'' attribution requirement do not apply to inference-only academic evaluation.
NVIDIA Nemotron Nano~2 \cite{nvidia2025nvidianemotronnano2} is released under the NVIDIA Open Model License, a custom permissive license allowing academic and commercial use. Attribution is required only when redistributing model weights, which we do not.

Our use of all evaluated models is inference-only academic benchmarking, consistent with the intended use specified by each license.
GPT-OSS additionally requires compliance with OpenAI's gpt-oss usage policy, which prohibits use for illegal or harmful content generation; our evaluation does not engage in any such use.

\subsection{Inference Setup}
\label{app:models-inference}

All inference is run with vLLM~0.20.0 (CUDA~12.9) on a workstation with four NVIDIA RTX Pro~6000 Blackwell GPUs. We run each model at its native released precision (e.g., BF16 for most checkpoints, MXFP4 for GPT-OSS). Maximum context length is 32{,}768 tokens and maximum generated length is 16{,}384 tokens per response. Generation parameters (temperature, top-$p$) are left at vLLM defaults; we do not tune decoding so that all models are compared under their default behaviour.

All belief elicitation calls use vLLM's \texttt{response\_format} with a JSON schema, constraining the output to a 7-point Likert label set. This avoids free-form parsing errors and keeps the output space identical across models.

For checkpoints with a built-in reasoning mode, we use vLLM's family-specific reasoning parser (\texttt{qwen3}, \texttt{gemma4}, \texttt{olmo3}, \texttt{glm45}, \texttt{openai\_gptoss}). These parsers capture the chain-of-thought in a separate channel, so it does not contaminate the final content. Reasoning is toggled per (checkpoint, reasoning-mode) configuration via the chat-template flag exposed by each family (e.g., \texttt{enable\_thinking} for Qwen, Gemma, and the Olmo \texttt{Think} variants; \texttt{reasoning\_effort} for GPT-OSS).

Each elicitation call is retried on failure, and we target a per-configuration completion rate of at least 95\%. Out of the 40 evaluated configurations, only 5 fall below 99\% completion; the remaining failures are caused either by reasoning chains exceeding the maximum generated length or by outputs that cannot be parsed as the required JSON schema.

Total compute cost (approximately 288 GPU-hours) is reported in the Ethics Statement.

\section{Dataset Details}
\label{app:dataset}

\subsection{Documentation}
\label{app:dataset-docs}

\paragraph{Language and Domain.}
All game dialogue, prompts, and belief responses are in English. The domain is a single multi-party social-deduction game, the seven-player variant of Werewolf, played in turn-based day/night cycles. The corpus contains no real-world text, no human-authored content, and no demographic data.

\paragraph{Data Generation.}
Each game is played by seven LLM agents drawn from a pool of seven open-weight models:
\begin{itemize}
    \item Qwen/Qwen3.6-35B-A3B
    \item Qwen/Qwen3.6-27B
    \item google/gemma-4-31B-it
    \item google/gemma-4-26B-A4B-it
    \item openai/gpt-oss-20b
    \item mistralai/Ministral-3-14B-Reasoning-2512
    \item zai-org/GLM-4.7-Flash
\end{itemize}
Reasoning is enabled for all models that support it. This game-generating pool is distinct from the 29 evaluated checkpoints in Appendix~\ref{app:models}. For each game, each agent is randomly assigned a role (two wolves, one seer, one doctor, three villagers). At every turn, an agent receives the game rules and the events observable to its assigned role, and is prompted to produce the next legal action. Games are played to completion before any belief elicitation begins, so the elicitation step cannot influence the gameplay or its outcome.

\paragraph{Annotation.}
Daytime messages are annotated by Claude Opus~4.6 into two categories. A \emph{suspect} expresses uncertainty about whether a target may be wolf-aligned; an \emph{accuse} directly states that a target is on the wolf team. For each annotated message we record the accuser, the accused target, and the role and team of each. A 150-message stratified subset was independently re-labelled by two of the authors to verify annotation reliability; the verification protocol and the 96\% agreement rate are reported in the Ethics Statement (``Annotation quality and annotator bias'').

\subsection{Statistics}
\label{app:dataset-stats}

Corpus size (200 games, 1{,}224 annotated messages) is reported in Section~\ref{sec:framework-data}. This work does not train or fine-tune any model, so all 1{,}224 messages are used as a single evaluation set with no train/validation/test partition.

\subsection{Result Reporting}
\label{app:result-reporting}

Tables~\ref{tab:exp2_belief_shift_stacked_rowlevel_with_ci} and \ref{tab:exp3_credulity_by_prior_by_size_stacked_rowlevel_with_ci} report the same belief shifts as the main result tables.
Each cell shows the mean with a 95\% confidence interval as a subscript.

\begin{table*}[t]
  \centering
  \footnotesize
  \setlength{\tabcolsep}{3.5pt}
  \renewcommand{\arraystretch}{0.95}
  \begin{tabular}{l r rr rr}
    \toprule
    & & \multicolumn{2}{c}{Target} & \multicolumn{2}{c}{Accuser} \\
    \cmidrule(lr){3-4} \cmidrule(lr){5-6}
    Size group & $n$ & Sus & Acc & Sus & Acc \\
    \midrule
    \multicolumn{6}{l}{\textbf{Villager as accuser}} \\
    $\le$4B & 12 & $0.24_{\pm 0.05}$ & $0.94_{\pm 0.03}$ & $-0.63_{\pm 0.04}$ & $-0.54_{\pm 0.03}$ \\
    5--9B & 8 & $0.52_{\pm 0.06}$ & $1.03_{\pm 0.05}$ & $-0.87_{\pm 0.06}$ & $-0.75_{\pm 0.04}$ \\
    10--29B & 9 & $0.64_{\pm 0.04}$ & $1.04_{\pm 0.03}$ & $-0.37_{\pm 0.04}$ & $-0.36_{\pm 0.03}$ \\
    $\ge$30B & 11 & $0.62_{\pm 0.03}$ & $0.96_{\pm 0.03}$ & $-0.47_{\pm 0.04}$ & $-0.44_{\pm 0.03}$ \\
    \cmidrule(lr){1-6}
    Avg. & 40 & $0.49_{\pm 0.02}$ & $0.98_{\pm 0.02}$ & $-0.57_{\pm 0.02}$ & $-0.51_{\pm 0.02}$ \\
    \midrule
    \multicolumn{6}{l}{\textbf{Wolf as accuser}} \\
    $\le$4B & 12 & $0.21_{\pm 0.04}$ & $0.70_{\pm 0.04}$ & $-0.64_{\pm 0.04}$ & $-0.47_{\pm 0.04}$ \\
    5--9B & 8 & $0.33_{\pm 0.06}$ & $0.88_{\pm 0.05}$ & $-0.77_{\pm 0.06}$ & $-0.73_{\pm 0.05}$ \\
    10--29B & 9 & $0.35_{\pm 0.04}$ & $0.78_{\pm 0.03}$ & $-0.09_{\pm 0.04}$ & $-0.14_{\pm 0.03}$ \\
    $\ge$30B & 11 & $0.34_{\pm 0.03}$ & $0.66_{\pm 0.03}$ & $-0.13_{\pm 0.04}$ & $-0.08_{\pm 0.03}$ \\
    \cmidrule(lr){1-6}
    Avg. & 40 & $0.30_{\pm 0.02}$ & $0.74_{\pm 0.02}$ & $-0.40_{\pm 0.02}$ & $-0.34_{\pm 0.02}$ \\
    \bottomrule
  \end{tabular}
  \caption{%
    Same as Table~\ref{tab:exp2_belief_shift_stacked}, with 95\% confidence intervals.
  }
  \label{tab:exp2_belief_shift_stacked_rowlevel_with_ci}
\end{table*}

\begin{table*}[t]
  \centering
  \footnotesize
  \setlength{\tabcolsep}{3pt}
  \renewcommand{\arraystretch}{0.95}
  \begin{tabular}{l r rrr rrr}
    \toprule
    & & \multicolumn{3}{c}{Target} & \multicolumn{3}{c}{Accuser} \\
    \cmidrule(lr){3-5} \cmidrule(lr){6-8}
    Size & $n$ & Trust & Neutral & Distrust & Trust & Neutral & Distrust \\
    \midrule
    \multicolumn{8}{l}{\textbf{Villager as accuser}} \\
    $\le$4B & 12 & $0.78_{\pm 0.04}$ & $0.72_{\pm 0.05}$ & $0.37_{\pm 0.08}$ & $-0.55_{\pm 0.03}$ & $-0.46_{\pm 0.04}$ & $-0.97_{\pm 0.09}$ \\
    5--9B & 8 & $1.04_{\pm 0.04}$ & $0.47_{\pm 0.08}$ & $0.49_{\pm 0.15}$ & $-0.75_{\pm 0.04}$ & $-0.84_{\pm 0.07}$ & $-0.99_{\pm 0.16}$ \\
    10--29B & 9 & $0.99_{\pm 0.03}$ & $0.82_{\pm 0.04}$ & $0.20_{\pm 0.12}$ & $-0.35_{\pm 0.02}$ & $-0.48_{\pm 0.05}$ & $0.11_{\pm 0.15}$ \\
    $\ge$30B & 11 & $0.96_{\pm 0.03}$ & $0.72_{\pm 0.03}$ & $0.10_{\pm 0.12}$ & $-0.43_{\pm 0.02}$ & $-0.63_{\pm 0.05}$ & $0.39_{\pm 0.15}$ \\
    \cmidrule(lr){1-8}
    Avg. & 40 & $0.94_{\pm 0.02}$ & $0.71_{\pm 0.02}$ & $0.33_{\pm 0.06}$ & $-0.51_{\pm 0.01}$ & $-0.57_{\pm 0.02}$ & $-0.56_{\pm 0.07}$ \\
    \midrule
    \multicolumn{8}{l}{\textbf{Wolf as accuser}} \\
    $\le$4B & 12 & $0.72_{\pm 0.05}$ & $0.50_{\pm 0.05}$ & $0.30_{\pm 0.06}$ & $-0.58_{\pm 0.05}$ & $-0.48_{\pm 0.04}$ & $-0.56_{\pm 0.06}$ \\
    5--9B & 8 & $1.18_{\pm 0.06}$ & $0.37_{\pm 0.07}$ & $0.26_{\pm 0.07}$ & $-0.73_{\pm 0.05}$ & $-0.80_{\pm 0.07}$ & $-0.71_{\pm 0.08}$ \\
    10--29B & 9 & $0.97_{\pm 0.04}$ & $0.62_{\pm 0.04}$ & $0.15_{\pm 0.05}$ & $-0.32_{\pm 0.04}$ & $-0.17_{\pm 0.05}$ & $0.18_{\pm 0.05}$ \\
    $\ge$30B & 11 & $1.00_{\pm 0.04}$ & $0.60_{\pm 0.03}$ & $-0.12_{\pm 0.04}$ & $-0.43_{\pm 0.04}$ & $-0.25_{\pm 0.05}$ & $0.47_{\pm 0.05}$ \\
    \cmidrule(lr){1-8}
    Avg. & 40 & $0.96_{\pm 0.02}$ & $0.53_{\pm 0.02}$ & $0.14_{\pm 0.03}$ & $-0.52_{\pm 0.02}$ & $-0.40_{\pm 0.02}$ & $-0.14_{\pm 0.03}$ \\
    \bottomrule
  \end{tabular}
  \caption{%
    Same as Table~\ref{tab:exp3_credulity_by_prior_by_size_stacked_rowlevel}, with 95\% confidence intervals.
  }
  \label{tab:exp3_credulity_by_prior_by_size_stacked_rowlevel_with_ci}
\end{table*}

\subsection{Frontier Closed-Source Models}
\label{app:frontier}

To investigate the results in frontier models, we conduct a preliminary experiment by evaluating six frontier closed-source models, including Claude Opus 4.8, Claude Haiku 4.5, GPT-5.6-sol, GPT-5.6-luna, Gemini 3.1 Pro, and Gemini 3.5 Flash.
For prior-role discrimination, it continues to improve for frontier models, from 1.31 for the $\ge$30B group to 2.01 for the frontier group.
Interestingly, the original pattern shown in Table~\ref{tab:exp2_belief_shift_with_frontier} does not continue at the frontier.
Frontier models discount accusations from wolf-side accusers and become more suspicious of the accusers themselves (0.53 and 0.50).
However, for belief shifts by prior trust experiment as shown in Table~\ref{tab:exp3_credulity_by_prior_with_frontier}, frontier models still follow trusted accusers (0.63 and 0.81), but with smaller shifts, while accusations from distrusted accusers produce negative target shifts ($-0.46$ and $-0.62$).
This preliminary experiment reveals that larger models are more resistant to distrusted sources, and an in-depth analysis of belief shifts in frontier models would be worthwhile.

\begin{table}[t]
  \centering
  \footnotesize
  \setlength{\tabcolsep}{4.5pt}
  \renewcommand{\arraystretch}{0.95}
  \begin{tabular}{l r rr rr}
    \toprule
    & & \multicolumn{2}{c}{Target} & \multicolumn{2}{c}{Accuser} \\
    \cmidrule(lr){3-4} \cmidrule(lr){5-6}
    Size group & $n$ & Sus. & Acc. & Sus. & Acc. \\
    \midrule
    \multicolumn{6}{l}{\textbf{Villager as accuser}} \\
    $\le$4B & 12 & $0.24$ & $0.94$ & $-0.63$ & $-0.54$ \\
    5--9B & 8 & $0.52$ & $1.03$ & $-0.87$ & $-0.75$ \\
    10--29B & 9 & $0.64$ & $1.04$ & $-0.37$ & $-0.36$ \\
    $\ge$30B & 11 & $0.62$ & $0.96$ & $-0.47$ & $-0.44$ \\
    \cmidrule(lr){1-6}
    Avg. & 40 & $0.49$ & $0.98$ & $-0.57$ & $-0.51$ \\
    \cmidrule(lr){1-6}
    Frontier & 6 & $0.34$ & $0.52$ & $0.00$ & $-0.26$ \\
    \midrule
    \multicolumn{6}{l}{\textbf{Wolf as accuser}} \\
    $\le$4B & 12 & $0.21$ & $0.70$ & $-0.64$ & $-0.47$ \\
    5--9B & 8 & $0.33$ & $0.88$ & $-0.77$ & $-0.73$ \\
    10--29B & 9 & $0.35$ & $0.78$ & $-0.09$ & $-0.14$ \\
    $\ge$30B & 11 & $0.34$ & $0.66$ & $-0.13$ & $-0.08$ \\
    \cmidrule(lr){1-6}
    Avg. & 40 & $0.30$ & $0.74$ & $-0.40$ & $-0.34$ \\
    \cmidrule(lr){1-6}
    Frontier & 6 & $-0.10$ & $-0.13$ & $0.53$ & $0.50$ \\
    \bottomrule
  \end{tabular}
  \caption{%
    Same as Table~\ref{tab:exp2_belief_shift_stacked}, with the six frontier closed-source models added as a separate group.
    \textbf{Avg.} is the average over the 40 open-weight configurations and does not include the frontier group.
  }
  \label{tab:exp2_belief_shift_with_frontier}
\end{table}

\begin{table*}[t]
  \centering
  \footnotesize
  \setlength{\tabcolsep}{3pt}
  \renewcommand{\arraystretch}{0.95}
  \begin{tabular}{l r rrr rrr}
    \toprule
    & & \multicolumn{3}{c}{Target} & \multicolumn{3}{c}{Accuser} \\
    \cmidrule(lr){3-5} \cmidrule(lr){6-8}
    Size & $n$ & Trust & Neutral & Distrust & Trust & Neutral & Distrust \\
    \midrule
    \multicolumn{8}{l}{\textbf{Villager as accuser}} \\
    $\le$4B & 12 & $0.78$ & $0.72$ & $0.37$ & $-0.55$ & $-0.46$ & $-0.97$ \\
    5--9B & 8 & $1.04$ & $0.47$ & $0.49$ & $-0.75$ & $-0.84$ & $-0.99$ \\
    10--29B & 9 & $0.99$ & $0.82$ & $0.20$ & $-0.35$ & $-0.48$ & $0.11$ \\
    $\ge$30B & 11 & $0.96$ & $0.72$ & $0.10$ & $-0.43$ & $-0.63$ & $0.39$ \\
    \cmidrule(lr){1-8}
    Avg. & 40 & $0.94$ & $0.71$ & $0.33$ & $-0.51$ & $-0.57$ & $-0.56$ \\
    \cmidrule(lr){1-8}
    Frontier & 6 & $0.63$ & $0.35$ & $-0.46$ & $-0.39$ & $-0.08$ & $1.14$ \\
    \midrule
    \multicolumn{8}{l}{\textbf{Wolf as accuser}} \\
    $\le$4B & 12 & $0.72$ & $0.50$ & $0.30$ & $-0.58$ & $-0.48$ & $-0.56$ \\
    5--9B & 8 & $1.18$ & $0.37$ & $0.26$ & $-0.73$ & $-0.80$ & $-0.71$ \\
    10--29B & 9 & $0.97$ & $0.62$ & $0.15$ & $-0.32$ & $-0.17$ & $0.18$ \\
    $\ge$30B & 11 & $1.00$ & $0.60$ & $-0.12$ & $-0.43$ & $-0.25$ & $0.47$ \\
    \cmidrule(lr){1-8}
    Avg. & 40 & $0.96$ & $0.53$ & $0.14$ & $-0.52$ & $-0.40$ & $-0.14$ \\
    \cmidrule(lr){1-8}
    Frontier & 6 & $0.81$ & $0.05$ & $-0.62$ & $-0.55$ & $0.43$ & $0.87$ \\
    \bottomrule
  \end{tabular}
  \caption{%
    Same as Table~\ref{tab:exp3_credulity_by_prior_by_size_stacked_rowlevel}, with the six frontier closed-source models added as a separate group.
    \textbf{Avg.} is the average over the 40 open-weight configurations and does not include the frontier group.
  }
  \label{tab:exp3_credulity_by_prior_with_frontier}
\end{table*}

\subsection{Belief Shifts by Accusation Correctness}
\label{app:accusation-correctness}

Table~\ref{tab:exp3_credulity_by_accusation_correctness} separates the results of Table~\ref{tab:exp3_credulity_by_prior_by_size_stacked_rowlevel} according to the true role of the accused target.
An accusation is true when the target is wolf-aligned, and false when the target is village-aligned.
The main findings remain consistent: (a) when the accuser is trusted, true and false accusations produce similarly large target shifts, (b) larger models show greater resistance to distrusted wolf accusers in both the true- and false-accusation groups.
These results indicate that the trend is not caused by mixing the two types of accusations.

\begin{table*}[t]
  \centering
  \scriptsize
  \setlength{\tabcolsep}{2pt}
  \renewcommand{\arraystretch}{1.05}
  \resizebox{\textwidth}{!}{%
  \begin{tabular}{l r r@{\hspace{2.5pt}}l r@{\hspace{2.5pt}}l r@{\hspace{2.5pt}}l r@{\hspace{2.5pt}}l r@{\hspace{2.5pt}}l r@{\hspace{2.5pt}}l}
    \toprule
    & & \multicolumn{6}{c}{Target} & \multicolumn{6}{c}{Accuser} \\
    \cmidrule(lr){3-8} \cmidrule(lr){9-14}
    Size & $n$ & \multicolumn{2}{c}{Trust} & \multicolumn{2}{c}{Neutral} & \multicolumn{2}{c}{Distrust} & \multicolumn{2}{c}{Trust} & \multicolumn{2}{c}{Neutral} & \multicolumn{2}{c}{Distrust} \\
    \midrule
    \multicolumn{14}{l}{\textbf{Villager as accuser}} \\
    $\le$4B & 12 & $0.78$ & $(0.82,\,0.68)$ & $0.72$ & $(0.86,\,0.48)$ & $0.37$ & $(0.39,\,0.34)$ & $-0.55$ & $(-0.59,\,-0.48)$ & $-0.46$ & $(-0.44,\,-0.48)$ & $-0.97$ & $(-0.90,\,-1.08)$ \\
    5--9B & 8 & $1.04$ & $(1.04,\,1.05)$ & $0.47$ & $(0.55,\,0.33)$ & $0.49$ & $(0.48,\,0.52)$ & $-0.75$ & $(-0.77,\,-0.71)$ & $-0.84$ & $(-0.88,\,-0.80)$ & $-0.99$ & $(-0.93,\,-1.07)$ \\
    10--29B & 9 & $0.99$ & $(0.98,\,1.03)$ & $0.82$ & $(0.95,\,0.61)$ & $0.20$ & $(0.13,\,0.33)$ & $-0.35$ & $(-0.35,\,-0.34)$ & $-0.48$ & $(-0.61,\,-0.32)$ & $0.11$ & $(0.26,\,-0.03)$ \\
    $\ge$30B & 11 & $0.96$ & $(0.92,\,1.08)$ & $0.72$ & $(0.81,\,0.58)$ & $0.10$ & $(0.11,\,0.08)$ & $-0.43$ & $(-0.42,\,-0.44)$ & $-0.63$ & $(-0.76,\,-0.45)$ & $0.39$ & $(0.50,\,0.28)$ \\
    \cmidrule(lr){1-14}
    Avg. & 40 & $0.94$ & $(0.94,\,0.96)$ & $0.71$ & $(0.82,\,0.51)$ & $0.33$ & $(0.33,\,0.33)$ & $-0.51$ & $(-0.52,\,-0.49)$ & $-0.57$ & $(-0.63,\,-0.49)$ & $-0.56$ & $(-0.52,\,-0.61)$ \\
    \midrule
    \multicolumn{14}{l}{\textbf{Wolf as accuser}} \\
    $\le$4B & 12 & $0.72$ & $(0.88,\,0.70)$ & $0.50$ & $(0.78,\,0.47)$ & $0.30$ & $(0.42,\,0.28)$ & $-0.58$ & $(-0.59,\,-0.58)$ & $-0.48$ & $(-0.39,\,-0.49)$ & $-0.56$ & $(-0.03,\,-0.63)$ \\
    5--9B & 8 & $1.18$ & $(1.07,\,1.19)$ & $0.37$ & $(0.36,\,0.37)$ & $0.26$ & $(0.05,\,0.29)$ & $-0.73$ & $(-0.64,\,-0.74)$ & $-0.80$ & $(-0.72,\,-0.80)$ & $-0.71$ & $(-0.24,\,-0.77)$ \\
    10--29B & 9 & $0.97$ & $(0.82,\,0.99)$ & $0.62$ & $(0.64,\,0.62)$ & $0.15$ & $(-0.19,\,0.19)$ & $-0.32$ & $(-0.22,\,-0.34)$ & $-0.17$ & $(-0.31,\,-0.16)$ & $0.18$ & $(0.66,\,0.11)$ \\
    $\ge$30B & 11 & $1.00$ & $(0.81,\,1.02)$ & $0.60$ & $(0.66,\,0.59)$ & $-0.12$ & $(-0.29,\,-0.10)$ & $-0.43$ & $(-0.37,\,-0.44)$ & $-0.25$ & $(-0.49,\,-0.23)$ & $0.47$ & $(0.69,\,0.44)$ \\
    \cmidrule(lr){1-14}
    Avg. & 40 & $0.96$ & $(0.89,\,0.97)$ & $0.53$ & $(0.66,\,0.52)$ & $0.14$ & $(0.00,\,0.16)$ & $-0.52$ & $(-0.45,\,-0.52)$ & $-0.40$ & $(-0.45,\,-0.39)$ & $-0.14$ & $(0.29,\,-0.20)$ \\
    \bottomrule
  \end{tabular}%
  }
  \caption{%
    Same as Table~\ref{tab:exp3_credulity_by_prior_by_size_stacked_rowlevel}, separated by the true role of the accused target.
    Each cell shows the overall mean, followed in parentheses by the means for true accusations (wolf-aligned target) and false accusations (village-aligned target).
  }
  \label{tab:exp3_credulity_by_accusation_correctness}
\end{table*}

\subsection{Response Distributions Over the Seven-Level Scale}
\label{app:scale-usage}

To examine whether models use the full range of the scale, we analyze the response distributions for both prior beliefs and belief shifts across model-size groups.
The results shown in Table~\ref{tab:likert_response_distribution} show that all seven categories are used in every size group.
For prior beliefs, responses are broadly distributed across the scale.
For belief shifts, 0 is the most frequent response, but both extreme values ($-3$ and $+3$) are still used across all size groups.

\begin{table}[t]
  \centering
  \footnotesize
  \setlength{\tabcolsep}{3.5pt}
  \renewcommand{\arraystretch}{0.95}
  \begin{tabular}{l r rrrrrrr}
    \toprule
    Size group & $n$ & $-3$ & $-2$ & $-1$ & $0$ & $+1$ & $+2$ & $+3$ \\
    \midrule
    \multicolumn{9}{l}{\textbf{Prior belief}} \\
    $\le$4B & 12 & 14.0 & 19.7 & 14.2 & 12.9 & 10.4 & 17.6 & 11.3 \\
    5--9B & 8 & 18.5 & 23.9 & 6.2 & 11.8 & 6.5 & 23.0 & 10.0 \\
    10--29B & 9 & 13.8 & 24.8 & 11.2 & 11.8 & 9.7 & 19.6 & 9.1 \\
    $\ge$30B & 11 & 10.8 & 25.2 & 12.8 & 11.7 & 11.5 & 19.2 & 8.9 \\
    \midrule
    \multicolumn{9}{l}{\textbf{Belief shift}} \\
    $\le$4B & 12 & 15.3 & 6.1 & 9.1 & 32.7 & 13.7 & 15.0 & 8.1 \\
    5--9B & 8 & 18.7 & 7.4 & 7.4 & 32.6 & 5.1 & 16.3 & 12.5 \\
    10--29B & 9 & 4.7 & 5.4 & 10.4 & 45.4 & 13.7 & 15.2 & 5.2 \\
    $\ge$30B & 11 & 5.7 & 6.1 & 12.0 & 41.8 & 13.1 & 15.9 & 5.2 \\
    \bottomrule
  \end{tabular}
  \caption{%
    Distribution of responses (\%) over the seven-level scale, for prior beliefs and for belief shifts, by model size group.
    $n$ is the number of models in each size group.
  }
  \label{tab:likert_response_distribution}
\end{table}

\end{document}